\documentclass[10pt, a4paper]{article}

\usepackage{amsmath, amsfonts, amssymb}
\usepackage{booktabs}
\usepackage{subcaption}
\usepackage{enumitem}
\usepackage{mathtools}
\usepackage{float}
\usepackage{lipsum}
\usepackage{hyperref}
\usepackage{cleveref}

\usepackage{lrec-coling2024} 

\title{On the Relationship between Skill Neurons and Robustness in Prompt Tuning}

\name{Leon Ackermann, Xenia Ohmer} 

\address{University of Osnabrueck \\
         Osnabrueck, Germany \\
         $\lbrace$lackermann, xenia.ohmer$\rbrace$@uni-osnabrueck.de}

\abstract{
Prompt Tuning is a popular parameter-efficient finetuning method for pre-trained large language models (PLMs). Based on experiments with RoBERTa, it has been suggested that Prompt Tuning activates specific neurons in the transformer's feed-forward networks, that are highly predictive and selective for the given task. In this paper, we study the robustness of Prompt Tuning in relation to these ``skill neurons'', using RoBERTa and T5. We show that prompts tuned for a specific task are transferable to tasks of the same type but are not very robust to adversarial data. While prompts tuned for RoBERTa yield below-chance performance on adversarial data, prompts tuned for T5 are slightly more robust and retain above-chance performance in two out of three cases. At the same time, we replicate the finding that skill neurons exist in RoBERTa and further show that skill neurons also exist in T5. Interestingly, the skill neurons of T5 determined on non-adversarial data are also among the most predictive neurons on the adversarial data, which is not the case for RoBERTa. We conclude that higher adversarial robustness may be related to a model's ability to consistently activate the relevant skill neurons on adversarial data. 
 \\ \newline \Keywords{parameter-efficient finetuning, Prompt Tuning, adversarial robustness, skill neurons, interpretability} }

\begin{document}

\maketitleabstract

\section{Introduction}

Pretrained large language models (PLMs) comprise increasingly large numbers of parameters. 
For example, while Roberta-Large has ``only'' 355 million parameters \citep{roberta}, T5-XXL has 11 billion parameters \citep{t5}, and LLama-2 up to 70 billion \citep{touvron2023llama}.
Finetuning such models for downstream tasks is extremely expensive both in terms of computation and storage. 
Parameter-efficient finetuning (PEFT) methods have been developed as a solution to this problem.
These methods adapt PLMs to downstream tasks by finetuning only a small set of (additional) parameters.

Next to Low Rank Adaptation (LoRA) \citep{hu2022lora}, Prefix Tuning \citep{li-liang-2021-prefix}, and P-Tuning \citep{liu2021ptuning}, 
Prompt Tuning \citep{lester-etal-2021-power} is one of the state-of-the-art PEFT methods for PLMs \citep[see e.g.,][]{peft}. 
In Prompt Tuning, prompt tokens are prepended to the model input \textit{in the embedding space}, and only these prepended tokens are learned during finetuning while the actual model parameters are frozen. 
In experiments with various T5 model sizes, \citet{lester-etal-2021-power} showed that Prompt Tuning performance is on par with conventional finetuning for larger models. 
The authors further demonstrated that---next to reducing computational and storage requirements---Prompt Tuning has the advantage of being more robust to domain shifts, as adapting fewer parameters reduces the risk of overfitting.

To understand how Prompt Tuning works, researchers have started looking at its effects on PLM activations. 
In general, it is known that activations in the feed-forward networks (FFNs) of transformers \citep{transformers} can specialize to encode specific knowledge \citep{dai-etal-2022-knowledge} or concepts \citep{expert_neurons}. 
For Prompt Tuning, it has been shown that the overlap between the FFN neurons activated by different prompts can predict prompt transferability \citep{su-etal-2022-transferability}.
More recently, \citet{wang-etal-2022-finding-skill} showed that the activations of some FFN neurons are highly predictive of the task labels after Prompt Tuning. 
Analyses by the authors indicate that these ``skill neurons'' are task-specific, essential for task performance, and likely already generated during pretraining.

Our work extends ongoing research on robustness and skill neurons and establishes a connection between these two aspects.
We run experiments with RoBERTa \citep{roberta} and T5 \citep{t5} to capture differences between encoder-only and encoder-decoder architectures and utilize various benchmark datasets that cover a broad spectrum of NLP tasks.
For each dataset and model, we tune several prompts (using different seeds) and identify the associated skill neurons. 
Our main contributions are: 

\begin{enumerate}
    \item Consistent with previous research, we find that tuned prompts can be transferred to other datasets, including datasets involving domain shifts, provided these datasets pertain to the same type of task. 
    \item Using \texttt{AdversarialGLUE} \citep{wang2021adversarial}, we show that Prompt Tuning is not robust to adversarial data.
    \item \citeauthor{wang-etal-2022-finding-skill}'s (\citeyear{wang-etal-2022-finding-skill}) skill neuron analysis was limited to RoBERTa. We replicate their findings and further identify skill neurons in (the encoder of) T5.
    \item We establish a potential link between adversarial robustness and skill neurons. T5 exhibits greater robustness to adversarial data than RoBERTa. While T5's skill neurons on adversarial data are relatively consistent with its skill neurons on the corresponding non-adversarial data, this is not the case for RoBERTa.
\end{enumerate}
Our code is publicly available at \url{https://github.com/LeonAckermann/robust-neurons}.

In conclusion, our study offers additional evidence supporting the existence of skill neurons in PLMs. Although Prompt Tuning typically lacks adversarial robustness, our findings indicate that a model's robustness against adversarial attacks may depend on its ability to maintain task-relevant skill neurons on adversarial data.
Since skill neurons already emerge during pretraining and are thus independent of Prompt Tuning, our results are of general relevance for PLMs.

\section{Related Work}\label{sec:related_work}

Prompt Tuning is a PEFT method. In line with other work, we examine to what extent tuned prompts generalize to other (non-adversarial) datasets of the same task as well as to adversarial datasets. Furthermore, we use the tuned prompts to identify whether specific neurons in the models encode specific skills, by analyzing the models' FFN activations.

\paragraph{Parameter-efficient Finetuning.}\label{paragraph:parameter-efficient-finetuning}
All PEFT methods train only a few (additional) parameters to adapt PLMs to a certain task. 
They can be divided into \textit{adapter-based}  and \textit{prompt-based} methods.
Adapter-based methods insert small neural modules (adapters) into the transformer layers, which are tuned to the task. They were first used in Computer Vision \citep{rebuffi_2017} and then also became popular in NLP \citep{houlsby_2019}, especially in the form of low-rank adapter tuning \citep{hu2022lora}. Instead of inserting additional modules, prompt-based methods extend the original inputs with additional parameters. An important example is Prefix-Tuning \citep{li-liang-2021-prefix}, which prepends virtual tokens to each layer in the encoder stack, including the input (embedding) layer. Prompt Tuning \citep{lester-etal-2021-power} further simplifies that method by only adding tokens to the input layer. Compared to adapter-based methods, prompt-based methods tend to converge more slowly, and perform worse with smaller datasets and models \citep[e.g.][]{Ding2023}. On the other hand, they are easy to implement and require even fewer changes to the model.

\paragraph{Prompt Transferability.}\label{paragraph:prompt-transferability}
An additional advantage of Prompt Tuning is that tuned prompts may be reusable. 
\citet{lester-etal-2021-power} showed that Prompt Tuning is robust to domain shifts by evaluating prompts, tuned on specific question answering (QA) and paraphrase identification datasets, on other QA and paraphrase detection datasets. Prompt Tuning was more robust than traditional finetuning, with an increasing advantage for larger domain shifts. 
Using a larger set of tasks, \citet{su-etal-2022-transferability} confirmed that prompts learned with Prompt Tuning can be transferred effectively to similar tasks (within models), and further showed that they can also be transferred between models (within tasks). 
The authors examined various prompt similarity metrics as transferability indicators. Especially the overlapping rate of activated neurons in the transformer FFN layers was indicative of prompt transferability---more so than metrics based on prompt similarity in the embedding space.

\paragraph{Adversarial Robustness.}\label{paragraph-adversarial-robustness}
We are interested in whether Prompt Tuning is also robust to adversarial examples. 
Adversarial examples are datapoints that are misclassified even though they are only slightly---often imperceptibly---different from correctly classified examples. \citet{adversarial_examples} discovered that several machine learning models, including neural networks, are vulnerable to such examples and that the same examples tend to be adversarial for different models. Several studies have shown that PLMs are affected as well \citep[e.g.,][]{garg-ramakrishnan-2020-bae,wang2021adversarial,Jin_Jin_Zhou_Szolovits_2020,zhang-etal-2021-sparsifying,li-etal-2020-bert-attack}. In the text domain, adversarial examples are generated through perturbations that preserve semantic meaning. Perturbations can be applied at the word level (e.g. synonym replacement, typos) or the sentence level (e.g. paraphrasing, adding distracting text). They can be generated automatically or crafted by humans \citep[e.g.][]{naik-etal-2018-stress, ribeiro-etal-2020-beyond, nie-etal-2020-adversarial,jia-liang-2017-adversarial}. A total of 14 perturbation methods were applied to the multitask benchmark \texttt{GLUE} \citep{wang-etal-2018-glue} to generate \texttt{AdvGLUE} \citep{wang2021adversarial}.

\paragraph{Analyzing FFN Activations in PLMs.}\label{paragraph:activation-space-analysis}

There is a wide interest in understanding the inner workings of PLMs and transformer FFN layers are increasingly studied in this context. 
In particular, evidence is growing that FFN layers serve as memories that store factual and linguistic knowledge. 
For example, \citet{geva-etal-2021-transformer} found that FFNs function similarly to key-value memories, in that they detect certain text input patterns and map them to an output distribution over tokens; and that the detected patterns contain increasingly semantic information when progressing through the transformer layers. 
Furthermore, knowledge encoded in the FFNs seems to be highly localized. 
Specific neurons seem to encode specific concepts or pieces of information, and by modifying the activations of these neurons, the model's expression of the corresponding knowledge can be regulated \citep{dai-etal-2022-knowledge, expert_neurons, kformer}. 
\citet{wang-etal-2022-finding-skill} analyzed the activations in FFN layers when prepending task-specific continuous prompts to the input (learned through Prompt Tuning). Their findings suggest that certain neurons encode task-specific skills and that these skills, like factual knowledge, are already acquired during model pretraining.

\section{Methods}

This section introduces Prompt Tuning more formally (Section \ref{subsec:prompt_tuning}) and describes how the predictivities of individual neurons can be determined using tuned prompts (Section \ref{subsec:predictivity}). 
The neurons with the highest predictivity for a specific task are considered the model's skill neurons for that task.

\subsection{Prompt Tuning}\label{subsec:prompt_tuning}

The model embeds input sequence $X_{orig} = [\text{token }1,\text{token }2, \dots, \text{token }s]$ into $\mathbf{X}\in\mathbb{R}^{s\times h}$, where $h$ is the embedding dimension. Prompt Tuning prepends additional prompt tokens $\mathbf{P} = [\mathbf{p}_1, \dots, \mathbf{p}_p], \mathbf{p}_i \in \mathbb{R}^h$ to that input in the embedding space, such that the new model input is $(\mathbf{P}, \mathbf{X}) = [\mathbf{p}_1, \dots, \mathbf{p}_p, \mathbf{x}_1, \dots, \mathbf{x}_s]$, with $(\mathbf{P}, \mathbf{X}) \in \mathbb{R}^{(p+s)\times h}$. The continuous prompt tokens in the embedding space are treated as free model parameters and their values are learned via backpropagation during the training phase. All other model parameters are frozen. Thus, Prompt Tuning does not change any of the model's original weights, and only a few new parameters ($p \times h$) are learned per task.

\subsection{Skill Neurons}\label{subsec:predictivity}

Based on the method by \citet{wang-etal-2022-finding-skill}, skill neurons are identified as neurons in the FFNs of a transformer model whose activations are highly predictive of the task labels. Skill neurons are defined in relation to task-specific prompts, such as the ones generated through Prompt Tuning. They are calculated in three steps: 1) The \textit{baseline activation} for each neuron is calculated. 2) The \emph{predictivity} of each neuron is calculated, and 3) The consistently most predictive neurons are identified as \textit{skill neurons}. In the following, we describe how the skill neurons of one FFN (one layer) are determined using Prompt Tuning. The method is described for binary classification tasks, which we use in our analyses. 

\paragraph{Notation.} An FFN with activation function $f$
can formally be defined as
\begin{equation}
    \operatorname{FFN}(\mathbf{x})=f\left(\mathbf{x} \mathbf{K}^{\top}+\mathbf{b}_1\right) \mathbf{V}+\mathbf{b}_2 \;,
\end{equation}
where $\mathbf{x} \in \mathbb{R}^{h}$ is the embedding of an input token, $\mathbf{K}, \mathbf{V} \in \mathbb{R}^{f \times h}$ are weight matrices, and $\mathbf{b}_1, \mathbf{b}_2$ are biases. 
Given that the first linear transformation produces the activations $\mathbf{a}=f\left(\mathbf{x} \mathbf{K}^{\top}+\mathbf{b}_1\right)$, $a_i$ is considered the activation of the $i$-th neuron on input $\mathbf{x}$.

\paragraph{Baseline Activations.} Let the training set be defined as $D_{\text{train}} = \left\{\left(\mathbf{X}_1, y_1\right),\left(\mathbf{X}_2, y_2\right), \ldots,\left(\mathbf{X}_{|D|}, y_{|D|}\right)\right\}$, with $\mathbf{X}_i\in \mathbb{R}^{s \times h}$ (where $s$ is the input sequence length), and $y_i \in \lbrace0,1\rbrace$. Let $\mathbf{P}$ be the task prompt with $\mathbf{P}=[\mathbf{p}_1, \dots, \mathbf{p}_p], \mathbf{p}_i \in \mathbb{R}^h$. The baseline activation $a_{\text{bsl}}(\mathcal{N}, \mathbf{p}_i) \in \mathbb{R}$ is defined as the average activation of neuron $\mathcal{N}$ for prompt token $\mathbf{p}_i$ across the training data.  Let $a(\mathcal{N}, \mathbf{t}, \mathbf{X}_i)$ be the activation of neuron $\mathcal{N}$ for token embedding $\mathbf{t}$ given input $\mathbf{X}_i$. Then
\begin{equation}
    a_{\text{bsl}}(\mathcal{N}, \mathbf{p}_i) = 
    \frac{1}{\left|D_{\text{train}}\right|} 
    \sum_{\mathbf{X}_i \in D_{\text{train}}} a\big(\mathcal{N}, \mathbf{p}_i, (\mathbf{P},\mathbf{X}_i)\big) \;.
\end{equation}

\paragraph{Predictivities.} The accuracy of neuron $\mathcal{N}$ is calculated over the validation set $D_{\mathrm{dev}}$ with respect to the baseline activations calculated on the training set as

\begin{align}
    \operatorname{Acc}(\mathcal{N},\mathbf{p}_i) = \;\;\;\;\;\;\;\;\;\;\;\;\;\;\;\;\;\;\;\;\;\;\;\;\;\;\;\;\;\;\;\;\;\;\;\;\;\;\;\;\;\;\;\;\;\;\;\;\;\; \nonumber\\ 
    \frac{\sum_{(\mathbf{X}_i, y_i) \in D_{\mathrm{dev}}} 
    \mathbf{1}_{
    [
    \mathbf{1}_{
    [
    a(\mathcal{N}, \mathbf{p}_i, (\mathbf{P},\mathbf{X}_i))
    > 
    a_{\mathrm{bsl}}(\mathcal{N}, \mathbf{p}_i)
    ]}
    = y_i
    ]
    }}
    {|D_{\mathrm{dev}}|} \;, \label{eq:neuron_accuracies}
\end{align}
where $\mathbf{1}_{\text {[condition] }} \in\{0,1\}$ is the indicator function. In other words, the neuron's accuracy describes how often (on average) activations above or below the baseline activation co-occur with a positive or a zero label, respectively. Finally, to account for the fact that inhibitory neurons may also encode skills, the predictivity per neuron and prompt token is calculated as 
\begin{equation}
\operatorname{Pred}(\mathcal{N}, \mathbf{p}_i) = \max\big(\operatorname{Acc}(\mathcal{N}, \mathbf{p}_i), 1 - \operatorname{Acc}(\mathcal{N}, \mathbf{p}_i)\big)\;.
\end{equation}

\paragraph{Skill Neurons.} Given that a set of $k$ continuous prompts are trained $\mathcal{P} = \lbrace \mathbf{P}_1, \dots, \mathbf{P}_k\rbrace$ (with different seeds), the final predictivity of each neuron is given by
\begin{equation}
    \operatorname{Pred}(\mathcal{N})=\frac{1}{k} \sum_{\mathbf{P}_i \in \mathcal{P}} \max_{\mathbf{p}_j \in \mathbf{P}_i} \operatorname{Pred}(\mathcal{N},\mathbf{p}_j) \;.\label{eq:final_predictivity}
\end{equation}
When sorting the neurons in the model based on their predictivity, the most predictive neurons are considered the model's ``skill neurons'' for the given task.\footnote{We never determine a fixed set of skill neurons. Our analyses either involve \emph{all} predictivities, or we modify the activations of the top $k$\% predictive neurons for different values of $k$.}

\begin{figure*}[htb]
    \centering
    \begin{subfigure}[b]{0.49\textwidth}
        \includegraphics[width=\textwidth]{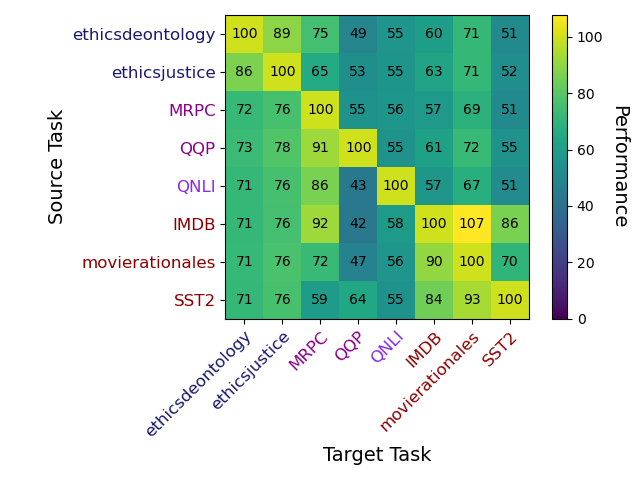} 
        \caption{RoBERTa}
        \label{fig:subfig1}
    \end{subfigure}
    \hfill
    \begin{subfigure}[b]{0.49\textwidth}
        \includegraphics[width=\textwidth]{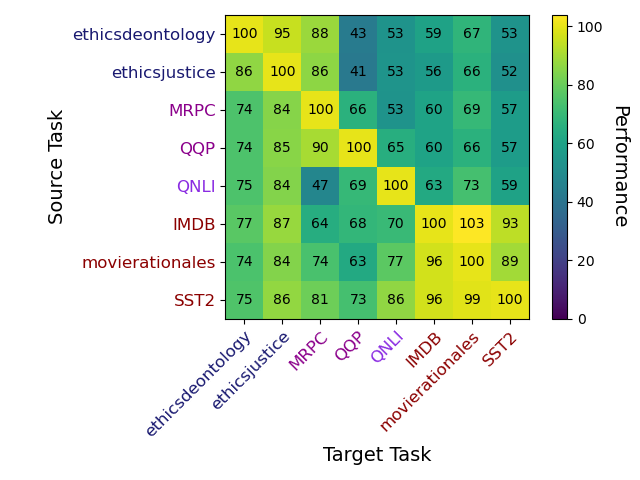} 
        \caption{T5}
        \label{fig:subfig2}
    \end{subfigure}
    \caption{Prompt transferability. We calculate the accuracy when using the prompt for the source task on the target task divided by the accuracy when using the prompt for the target task on the target task for each seed, and report the average across seeds.}
    \label{fig:transfer_performance}
\end{figure*}

\section{Experiments}

\paragraph{Models.} We run our experiments with RoBERTa-base (125M parameters) \citep{roberta} and T5-base (223M parameters) \citep{t5}.

\paragraph{Tasks.} 
We tune prompts for various types of binary classification tasks:
(1) paraphrase detection, including \texttt{QQP} \citep{wang-etal-2018-glue} and \texttt{MRPC} \citep{dolan-brockett-2005-automatically};
(2) sentiment analysis, including \texttt{Movie Rationales} \citep{zaidan-eisner-piatko-2008:nips}, \texttt{SST2} \citep{socher-etal-2013-recursive}, and \texttt{IMDB} \citep{maas-etal-2011-learning};
(3) ethical judgment, including \texttt{Ethics-Deontology} and \texttt{Ethics-Justice} \citep{hendrycks2020aligning},
and (4) natural language inference (NLI), including \texttt{QNLI} \citep{wang-etal-2018-glue}. We had also planned to include \texttt{RTE} (and \texttt{AdvRTE}) but performance after Prompt Tuning was poor, with accuracies between 55\%--60\% at a 50\% chance level.

Importantly, datasets belonging to the same task cover domain and format shifts. The paraphrase detection tasks consist of question pairs from \textit{Quora} (\texttt{QQP}) and sentence pairs from Newswire articles (\texttt{MRPC}). The sentiment analysis tasks are based on movie reviews from different websites and sometimes contain the full review (\texttt{IMDB}) and sometimes single sentences (\texttt{SST2}). The two ethics datasets are crowdsourced and focus on different types of ethical judgments, related to justice (\texttt{Ethics-Justice}) or deontology (\texttt{Ethics-Deontology}).

To test adversarial robustness we use \texttt{AdvQQP}, \texttt{AdvQNLI}, and \texttt{AdvSST2} from \texttt{AdvGLUE} \citep{wang2021adversarial}.
We work with the validation sets of the adversarial tasks since the submission format for evaluation on the test sets does not allow for a skill-neuron analysis.

\paragraph{Prompt Tuning.} We build on the code by \citet{su-etal-2022-transferability} and use the same parameters for Prompt Tuning.
In particular, the learned prompts consist of 100 (continuous) tokens.  
Their repository (\url{https://github.com/thunlp/Prompt-Transferability/}) includes one tuned prompt for each of the (non-adversarial) datasets that we use. 
We train four additional prompts (with different seeds) per dataset, resulting in a total of five prompts per dataset. 

\paragraph{Skill Neurons.} We calculate the neuron predictivities (Equation \ref{eq:final_predictivity}) for all non-adversarial datasets following the method described in Section \ref{subsec:predictivity}. 
We use the baseline activations from the non-adversarial datasets to calculate the neuron predictivities on the corresponding adversarial datasets. 
All analyses involving neuron predictivities are conducted simultaneously for each layer in the model, or each layer in the encoder model in the case of T5. The calculation of skill neurons relies on neuron activations for specific prompt tokens, which can be extracted from the encoder but not the decoder.

\begin{figure*}[htb]
    \centering
    \begin{subfigure}[b]{0.49\textwidth}
        \includegraphics[width=\textwidth]{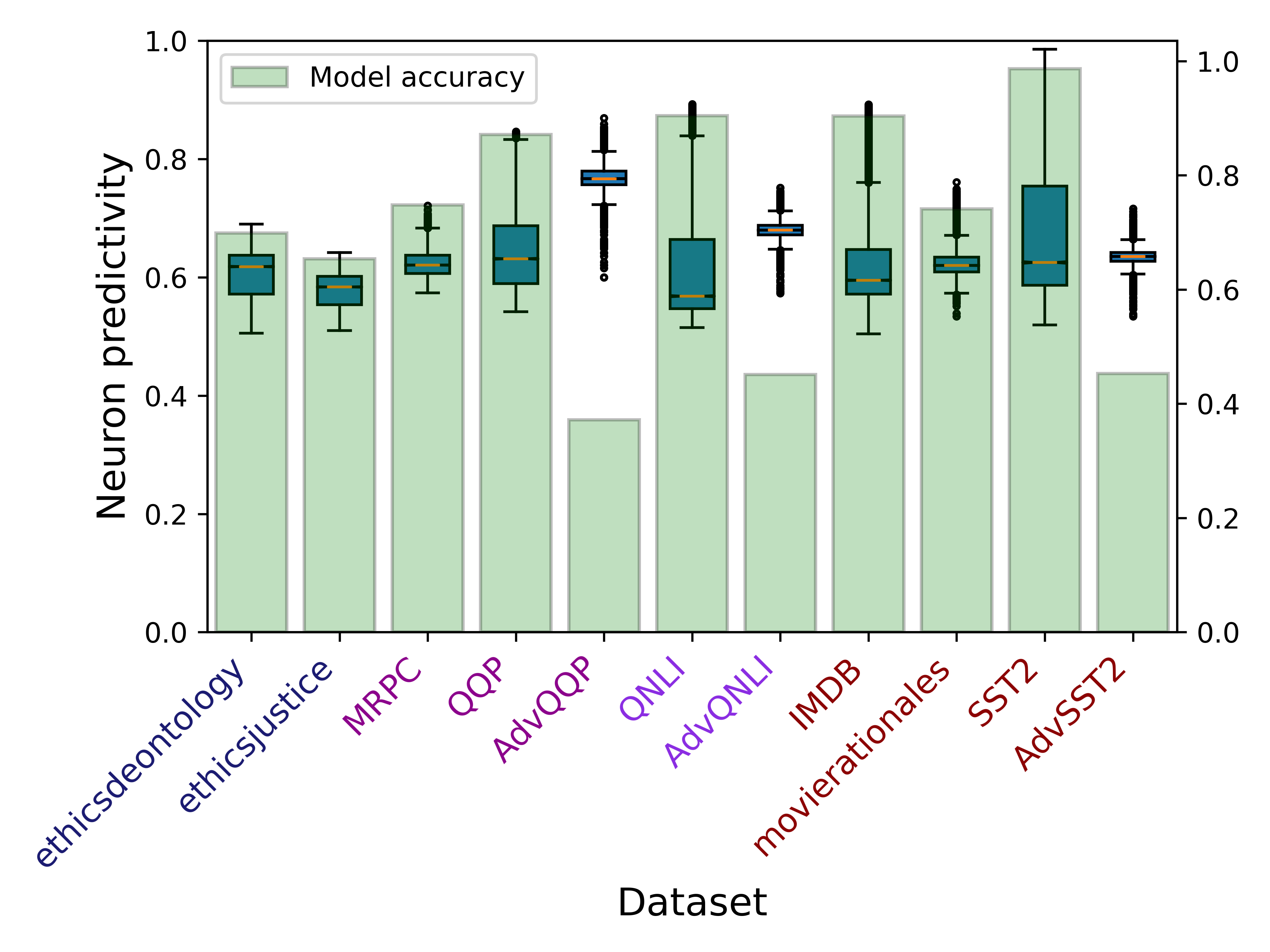}
        \caption{RoBERTa}
        \label{fig:results_boxplot_pred_roberta}
    \end{subfigure}
    \hfill 
    \begin{subfigure}[b]{0.49\textwidth}
        \includegraphics[width=\textwidth]{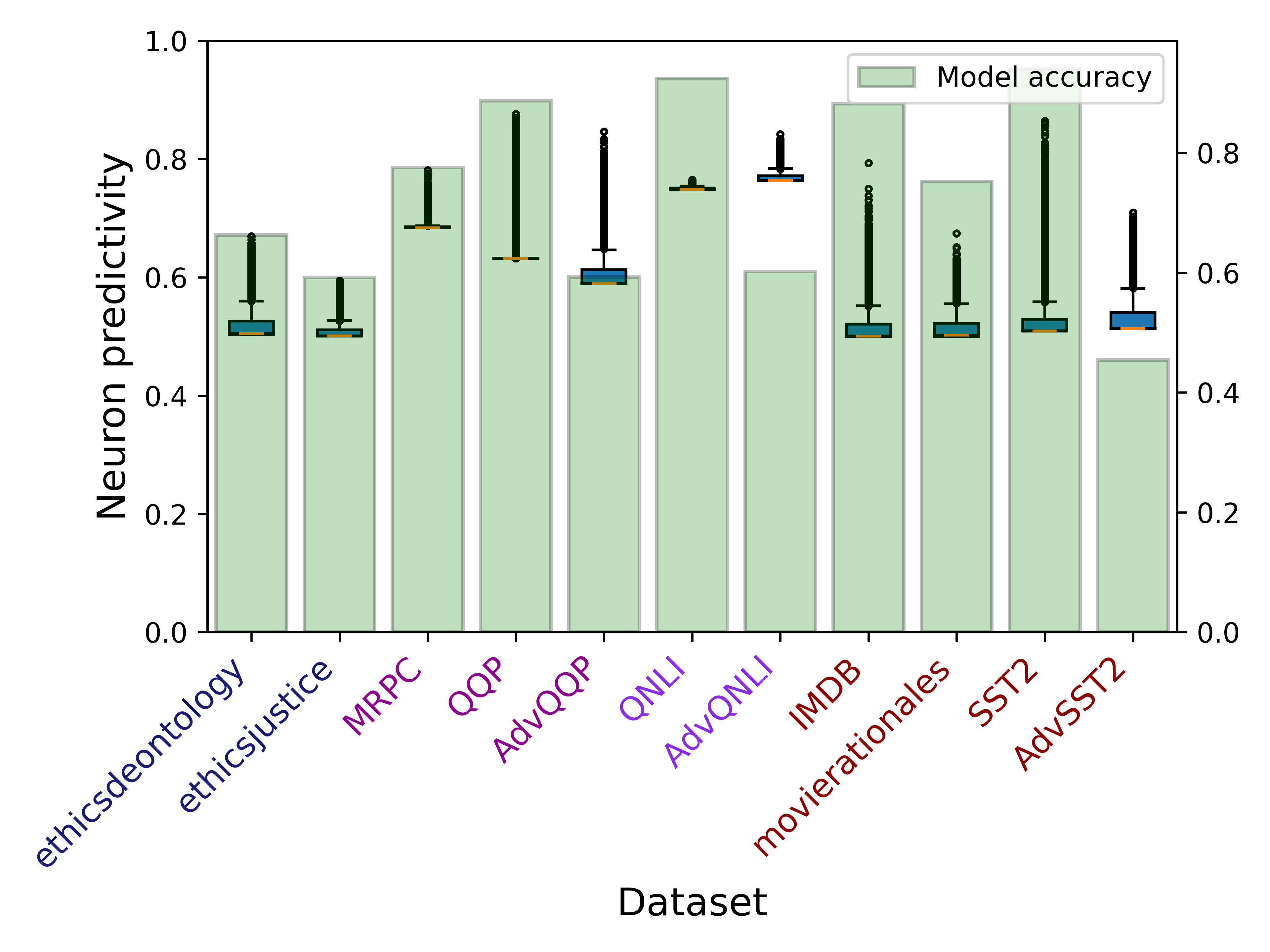}
        \caption{T5}
        \label{fig:results_boxplot_pred_t5}
    \end{subfigure}
    \caption{Distribution of neuron predictivities (box plots) on top of model accuracy (bar plots).}
    \label{fig:results_boxplot_pred_acc_per_layer}
\end{figure*}

\section{Results}

\subsection{Prompt Tuning and Robustness}

\paragraph{Prompt Tuning.}
We report mean accuracies and standard deviations across the five random seeds in Table \ref{tab:results_prompt_tuning_acc}.
Both the accuracies and the observed variations between seeds are in line with the results from other Prompt Tuning experiments \cite[e.g.][]{lester-etal-2021-power, su-etal-2022-transferability}. 
Performance is lowest on the ethical judgment tasks, with high accuracies on at least one dataset from the other tasks.
Overall, the performance of the two models is similar, with a slight advantage for RoBERTa on ethical judgment and sentiment analysis, and a slight advantage for T5 on paraphrase detection and NLI.

\begin{table}[htb]
\begin{tabular}{l|c|c}
\toprule
\textbf{Dataset} & \textbf{RoBERTa} & \textbf{T5} \\
\midrule
ethicsdeontology & $69.9 \pm 2.0$ & $66.3 \pm 1.6$ \\
ethicsjustice & $65.4 \pm 1.6$ & $59.1 \pm 2.9$ \\
\midrule
MRPC & $74.8 \pm 5.9$ & $77.5 \pm 2.6$ \\
QQP & $87.1 \pm 0,2$ & $88.7 \pm 1.1$ \\
AdvQQP & $37.2 \pm 4.1$ & $59.2 \pm 8.0$ \\
\midrule
QNLI & $90.4 \pm 0.2$ & $92.4 \pm 0.2$ \\
AdvQNLI & $45.1 \pm 3.5$ & $60.1 \pm 3.1$ \\
\midrule
IMDB & $90.4 \pm 0.3$ & $88.2 \pm 0.2$ \\
movierationales & $74.1 \pm 2.4$ & $75.2 \pm 1.4$ \\
SST2 & $98.7 \pm 2.6$ & $94.0 \pm 0.4$ \\
AdvSST2 & $45.3 \pm 4.5$ & $45.4 \pm 3.3$ \\
\bottomrule
\end{tabular}
\caption{Mean and standard deviation of the models' accuracy after Prompt Tuning.}
\label{tab:results_prompt_tuning_acc}
\end{table}

\paragraph{Robustness.}
We analyze two different kinds of robustness: adversarial robustness and transferability. 
Table \ref{tab:results_prompt_tuning_acc} shows the models' accuracy on the adversarial datasets, evaluated with the prompts of their non-adversarial counterparts.
The accuracies drop significantly. 
For RoBERTa, they are consistently below chance performance. The score is especially low for \texttt{AdvQQP}, which is unbalanced (41\% versus 59\%)---unlike \texttt{AdvQNLI} and \texttt{AdvSST2}.
T5 is somewhat more robust, with below chance performance on \texttt{AdvSST2} but around 60\% accuracy on the other two adversarial datasets.\footnote{F1 scores on \texttt{AdvQQP} range between 36.4--44.4\% for RoBERTa, with an average of 39.4\%; and between 38.9--59.7\% for T5, with an average of 48.2\%.}
Figure \ref{fig:transfer_performance} shows the relative task accuracies when transferring a continuous prompt from a source task to a target task (see Appendix \ref{app:transferability} for absolute values).
In line with earlier findings, the prompts tend to be highly transferable to datasets belonging to the same type of task \citep{lester-etal-2021-power, su-etal-2022-transferability}. 
In conclusion, Prompt Tuning is robust to data changes, including domain shifts (within the same type of task), but not to adversarial data.

\begin{figure*}[htb]
    \centering
    \begin{subfigure}[b]{0.49\textwidth}
        \includegraphics[width=\textwidth]{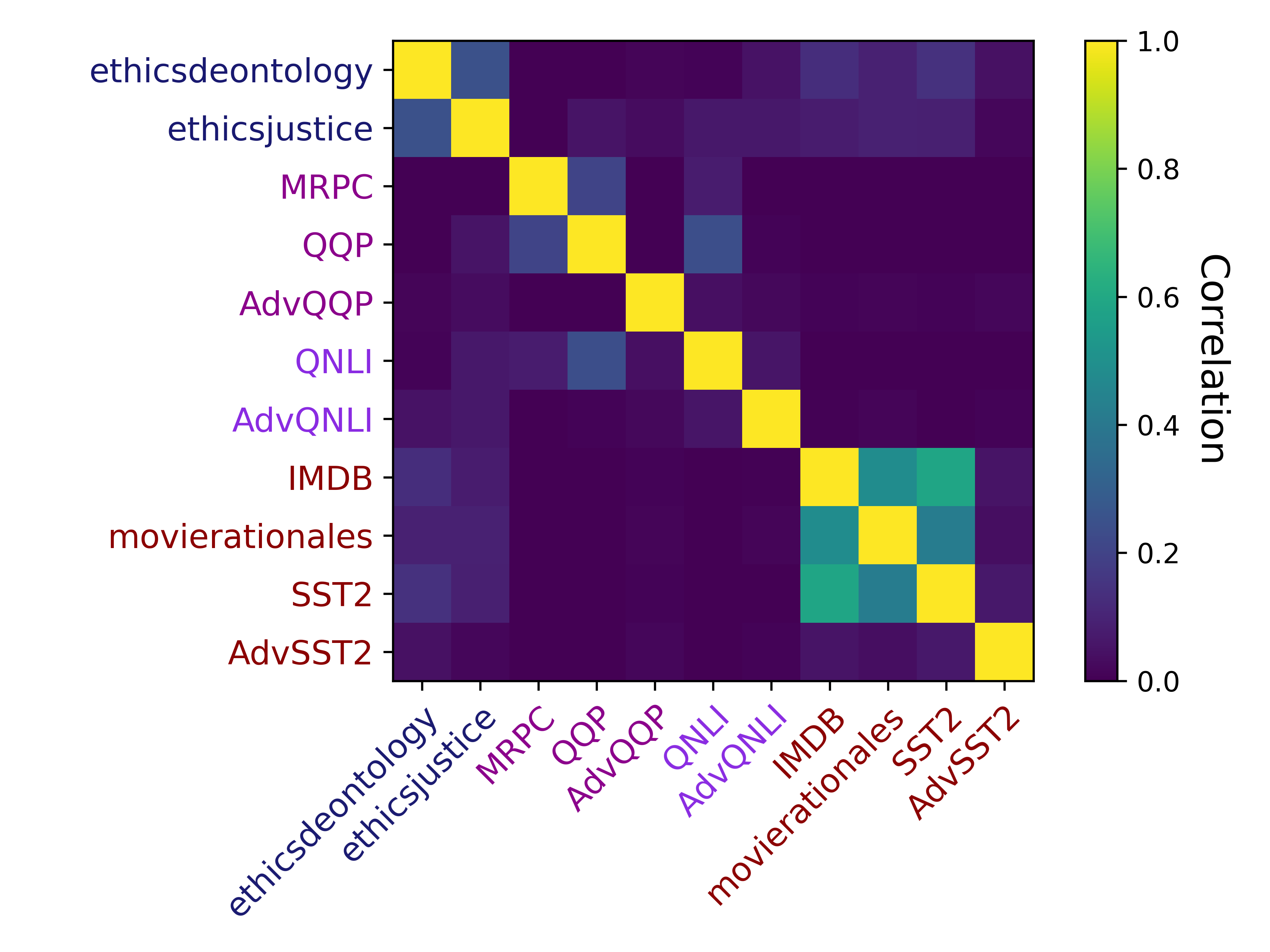}
        \caption{RoBERTa}
        \label{fig:results_task_specificity_roberta}
    \end{subfigure}
    \hfill 
    \begin{subfigure}[b]{0.49\textwidth}
        \includegraphics[width=\textwidth]{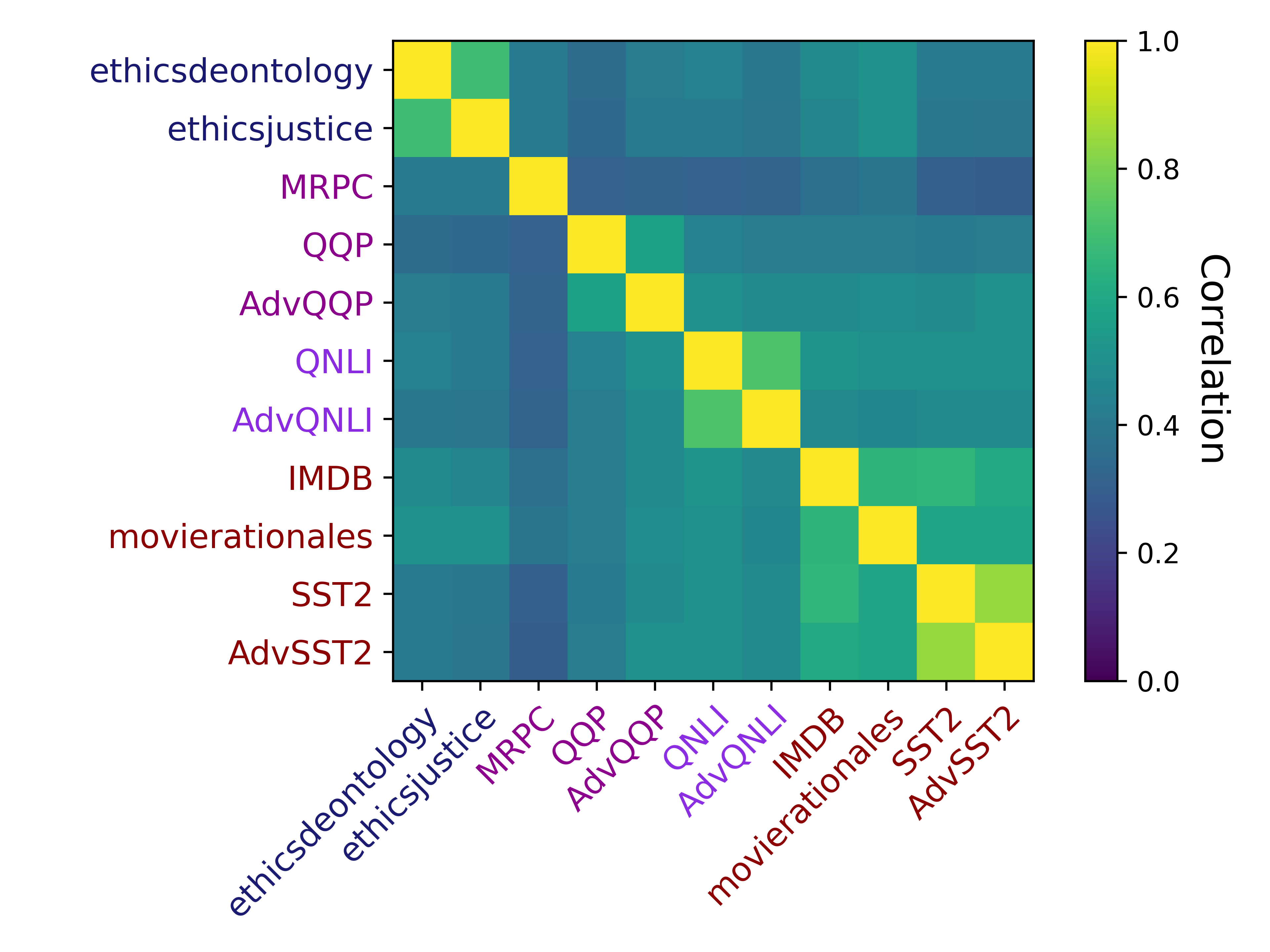}
        \caption{T5}
        \label{fig:results_task_specificity_t5}
    \end{subfigure}
    \caption{Spearman rank correlation between the neuron predictivities for different datasets.}
    \label{fig:results_task_specificity}
\end{figure*}

\subsection{Skill Neurons}\label{subsec:results_skill_neurons}

Following a similar procedure to \citet{wang-etal-2022-finding-skill}, we test for the existence of skill neurons by calculating the neuron predictivities (Equation \ref{eq:final_predictivity}) and making sure that the most predictive neurons are \textit{highly predictive}, \textit{task-specific}, and \textit{important} for solving the task.

\paragraph{High Predictivity.}
RoBERTa and T5 have only a few highly predictive neurons for each dataset.
This observation is in line with the initial findings presented by \citet{wang-etal-2022-finding-skill}. 
Consistent with other studies that analyzed FFN activations across layers \citep{geva-etal-2021-transformer, dai-etal-2022-knowledge}, neurons with high predictivities tend to be located in the upper layers. 
The predictivities of the most predictive neurons of RoBERTa largely correspond to the model's accuracy for the non-adversarial datasets (see Figure \ref{fig:results_boxplot_pred_roberta}). 
The most predictive neurons of T5 reach or fall (slightly) short of the model's accuracy (see Figure \ref{fig:results_boxplot_pred_t5}). 
For T5, especially in the cases where neuron predictivity is lower than model accuracy, more predictive neurons can probably be found in the decoder.
Regarding the adversarial datasets, the predictivities of almost all neurons exceed the models' accuracy.
Possible reasons are discussed in section \ref{subsec:relation_robustness_skill_neurons}. 

\paragraph{Task-specificity.} \label{para:task_specificity}
We calculate the Spearman rank correlation between the neuron predictivities for all datasets (see Figure \ref{fig:results_task_specificity}). 
The correlations are calculated per layer, based on the neuron predictivities when evaluated on the corresponding dataset, and then averaged across layers. 
High values within but not between different types of tasks for RoBERTa and T5 indicate a high task-specificity of the models' skill neurons. 
Notably, the correlations are generally higher for T5 which might be due to its sparse activations \citep{li_lazy_2022}. 
If there is a large number of consistently inactive neurons, they will always rank lower than active neurons and thus lead to a net positive correlation.

\paragraph{Importance.}
To ensure that the most predictive neurons are in fact essential for performing the task, 
we compare the decrease in accuracy when suppressing 1-15\% of the models' most predictive neurons versus the same number of random neurons. \citet{wang-etal-2022-finding-skill} perturb the neurons with Gaussian noise instead of suppressing them completely. 
Given that the activations in T5 are much higher than those in RoBERTa, using the same amount of noise (same standard deviation) will have different effects on the models. 
Instead, we decided to suppress the neurons, which has also been done in other work \citep[e.g.][]{dai-etal-2022-knowledge}.
Neurons are suppressed by setting their activations to zero.
For both models, the accuracy drops much more when suppressing skill neurons compared to random neurons, highlighting the importance of skill neurons for the models' task performance (see the \texttt{IMBD} example in Figure \ref{fig:suppression_analysis_imdb} and results for all datasets in Appendix \ref{app:suppression_analysis}).
In general, suppressing random neurons has a larger impact on RoBERTa than T5, which we again attribute to T5's sparse activations: A significant proportion of the randomly selected neurons would not have been active anyway. 

\begin{figure}[htb]
    \centering
    \includegraphics[width=0.45\textwidth]{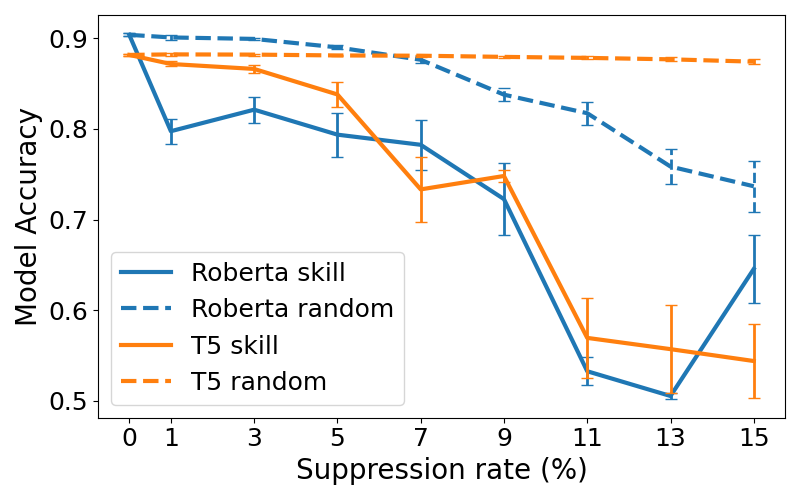}
    \caption{Model accuracies on \texttt{IMDB} when suppressing skill neurons (solid lines) versus randomly selected neurons (dashed lines).}
    \label{fig:suppression_analysis_imdb}
\end{figure}

In sum, both models have neurons that are predictive and selective for specific tasks, and their performance declines when suppressing these neurons, especially in comparison to suppressing random neurons.

 \begin{figure*}[htb]
    \centering
    \begin{subfigure}{0.45\textwidth}
        \centering
        \includegraphics[width=\textwidth]{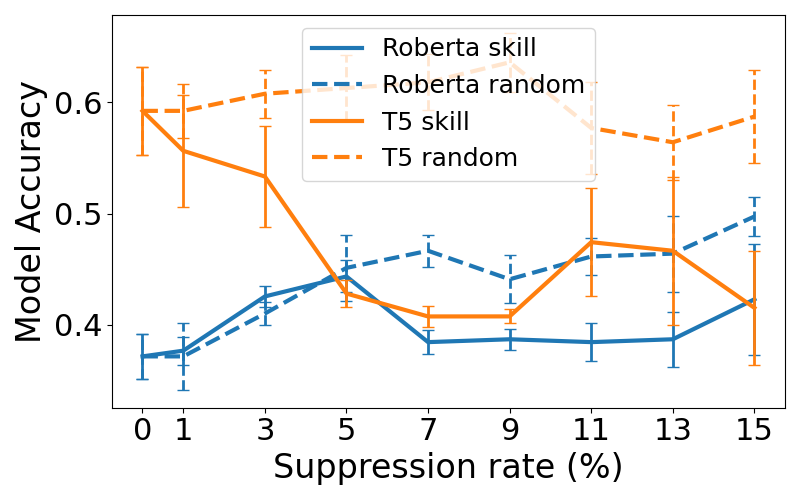}
        \caption{AdvQQP}
    \end{subfigure}
    \hfill
    \begin{subfigure}{0.45\textwidth}
        \centering
        \includegraphics[width=\textwidth]{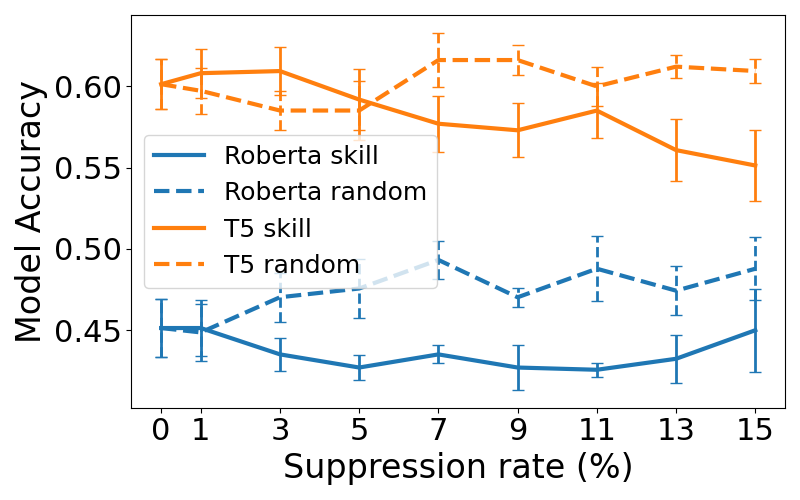}
        \caption{AdvQNLI}
    \end{subfigure}
    \caption{Model accuracies on each adversarial dataset when suppressing the skill neurons determined for these tasks (solid lines) and when suppressing randomly selected neurons (dashed lines).}
    \label{fig:suppression_analysis_adversarial}
\end{figure*}

\subsection{The Relationship between Robustness and Skill Neurons}
\label{subsec:relation_robustness_skill_neurons}

Our analyses above (Figure \ref{fig:results_boxplot_pred_acc_per_layer}) show that the most predictive neurons on the adversarial datasets are, in fact, more predictive than the model itself.
Notably, this is also true for the neuron accuracies (Equation \ref{eq:neuron_accuracies}) and can therefore not be attributed to inhibitory activations (see Appendix \ref{app:accuracies}).
These findings suggest that highly predictive neurons may exist that do not function as skill neurons because they do not encode the necessary skill (e.g. do not correlate with the skill neurons determined on similar tasks) or because the model does not rely on their activations in making a prediction.

To investigate these possibilities, we look at the Spearman rank correlation between the neuron predictivities on the adversarial datasets and the corresponding non-adversarial datasets (see Figure \ref{fig:results_task_specificity}). 
There are important differences between RoBERTa and T5. 
T5 exhibits strong ($\rho$: $0.57$--$0.84$) and significant ($p<0.01$) correlations between the predictivities. 
For RoBERTa, in contrast, correlations are close to zero ($\rho$: -$0.01$--$0.07$), and largely non-significant---except for \texttt{(Adversarial) QNLI} ($p=0.02$).
Even when accounting for the generally higher correlations for T5 (by normalizing the scores, see Appendix \ref{app:task_specificity}), T5 still exhibits a much stronger correspondence between adversarial and non-adversarial predictivities than RoBERTa.

Additionally, we study what happens when skill neurons for adversarial datasets are suppressed, ignoring the datasets where model performance is below chance to begin with, leaving us with T5: \texttt{AdvQQP} and \texttt{AdvQNLI}. Figure \ref{fig:suppression_analysis_adversarial} shows the model accuracies on these tasks. The results for RoBERTa are included for comparison. 
In both cases, suppressing the skill neurons leads to a decrease in performance, with a stronger decrease when more neurons are suppressed. Thus, the analyses so far establish that T5's ``adversarial'' skill neurons are important for the model's performance and further that they correlate with the ``non-adversarial'' skill neurons of the corresponding task. 

\begin{figure*}[htb]
    \centering
    \begin{subfigure}{0.45\textwidth}
        \centering
        \includegraphics[width=\textwidth]{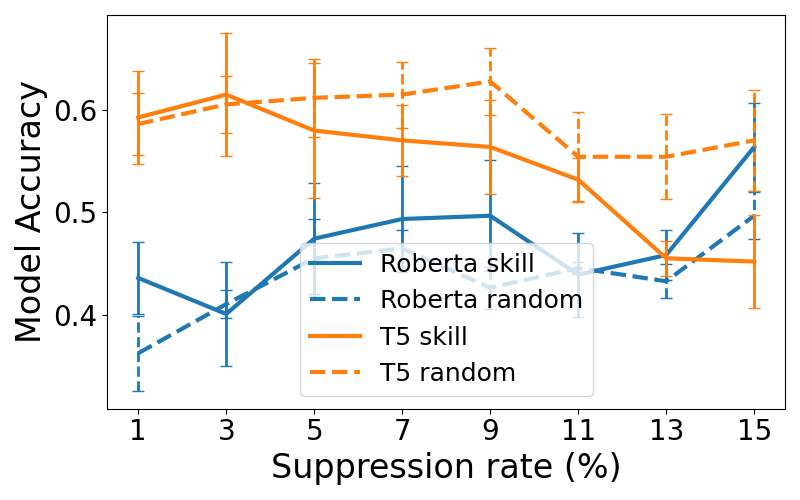}
        \caption{AdvQQP--QQP}
    \end{subfigure}
    \hfill
    \begin{subfigure}{0.45\textwidth}
        \centering
        \includegraphics[width=\textwidth]{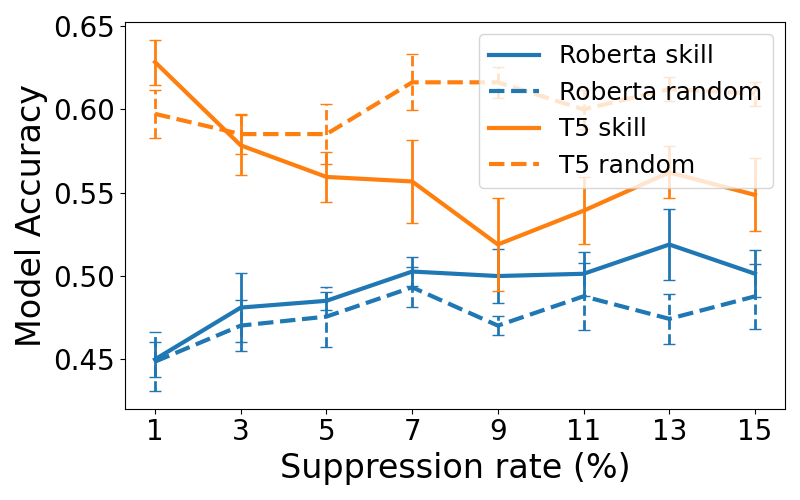}
        \caption{AdvQNLI--QNLI}
    \end{subfigure}
    \begin{subfigure}{0.45\textwidth}
        \centering
        \includegraphics[width=\textwidth]{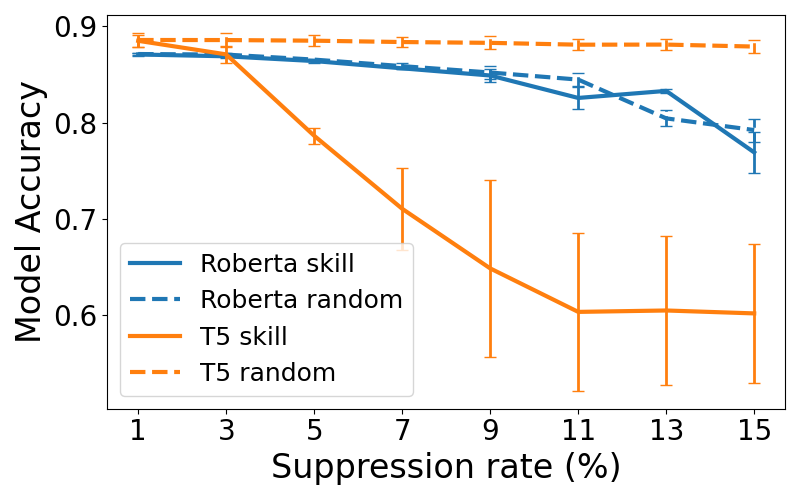}
        \caption{QQP--AdvQQP}
    \end{subfigure}
    \hfill
    \begin{subfigure}{0.45\textwidth}
        \centering
        \includegraphics[width=\textwidth]{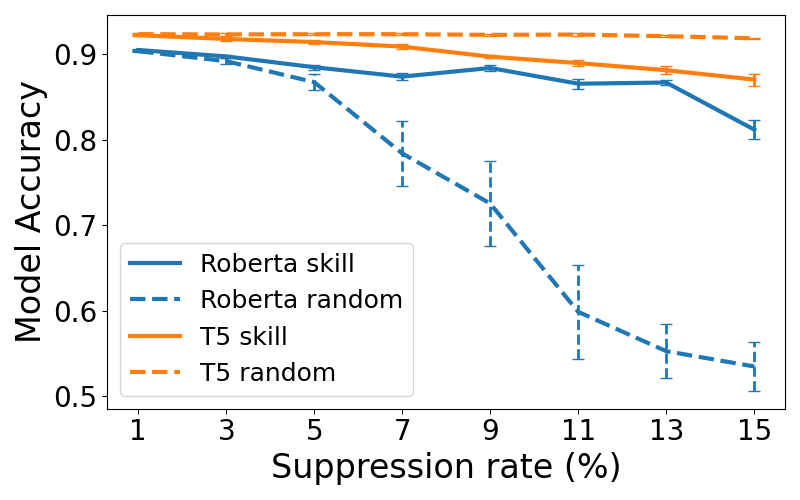}
        \caption{QNLI--AdvQNLI}
    \end{subfigure}
    \caption{Model accuracies on the adversarial datasets when suppressing the skill neurons identified on the corresponding non-adversarial datasets, and vice versa. For example, (a) shows the accuracies on \texttt{AdvQQP} when the most predictive neurons of \texttt{QQP} (solid lines) or randomly selected neurons (dashed lines) are suppressed.}
    \label{fig:suppression_analysis_adversarial_adv_nonAdv}
\end{figure*}

To further test whether T5 uses the same set of skill neurons on both adversarial and non-adversarial data, we run an ablation experiment: We evaluate the model's performance on the adversarial datasets when suppressing the skill neurons identified on the corresponding non-adversarial datasets and vice versa (see Figure \ref{fig:suppression_analysis_adversarial_adv_nonAdv}). 
Indeed, in both cases, performance is negatively affected, and suppressing the alternative skill neurons decreases performance more strongly than suppressing random neurons. 
For RoBERTa, in contrast, suppressing the alternative skill neurons is not more (and sometimes even less) harmful to performance than suppressing random neurons. In line with the correlation analysis, these results further support that at least some of T5's (but not RoBERTa's) skill neurons continue to be predictive and influential on adversarial data.
Taken together, these findings suggest that T5's higher robustness to adversarial data might be related to the fact that it recruits some of the skill neurons determined on the corresponding non-adversarial dataset, and therefore---given the high prompt transferability---neurons that generally encode knowledge about the relevant type of task.

\section{Discussion}
\label{sec:discussion}

This paper investigated the robustness of Prompt Tuning with respect to model activations. 

Firstly, we demonstrated that Prompt Tuning leads to a high prompt transferability between datasets of the same type of task but is not robust to adversarial data. 
Regarding adversarial robustness, T5 is more robust than RoBERTa, probably because the examples in \texttt{AdvGLUE} were generated against surrogate models based on BERT and RoBERTa \cite{devlin-etal-2019-bert}. 
Comparing our results on \texttt{AdvGLUE} to those of finetuned (and larger model versions of) T5 and RoBERTa \citep{wang2021adversarial} suggests that there is no advantage of Prompt Tuning over model finetuning in terms of adversarial robustness.
In other words, susceptibility to adversarial attacks likely arises during pretraining and persists across different task-adaptation methods. 

Secondly, we identified skill neurons in both RoBERTa and T5. 
The suppression analysis revealed that, even though both models rely on these skill neurons when performing a task, suppressing them affects RoBERTa more strongly than T5.
It might be that T5 encodes more redundant information. 
For example, it is known that a transformer's encoder output can be significantly compressed before being passed to the decoder without negatively impacting performance \citep{zhang-etal-2021-sparsifying}.
Besides, the skill neurons we identified for T5 tend to be slightly less predictive than those we identified for RoBERTa. 
More predictive neurons possibly reside in the T5 decoder and future work should extend the skill neuron analysis method to encompass both the transformer encoder and decoder. 

Finally, we identified a potential link between robustness and skill neurons. 
Our results suggest that the activation (as measured by the correlation analysis) and use (as measured by the suppression analysis) of the same skill neurons on non-adversarial and the corresponding adversarial data may be related to model robustness. 
We discussed above that the adversarial attacks in \texttt{AdvGLUE} were generated against BERT- and RoBERTa-based models, and that T5 is slightly more robust against these adversarial attacks. 
Given that T5, but not RoBERTa, use similar skill neurons on corresponding pairs of adversarial and non-adversarial datasets, it could be that adversarial attacks work because they modify relevant skill neuron activations.
Unlike RoBERTa, T5 has sparse activations.
While sparsity might explain some of our results, such as the task specificity and importance of the skill neurons (see \cref{subsec:relation_robustness_skill_neurons}), it is still an open question whether there is a connection between sparsity and adversarial robustness. 

Building on our insight that adversarial robustness may be regulated by individual neuron activations, future research on model robustness could aim to develop methods for consistently activating the relevant skill neurons for a given task.
For example, given that prompts are transferable between similar tasks, one could search for a prompt that activates skill neurons both on non-adversarial and adversarial data for a specific task, and transfer this prompt to other tasks of that type.
Similar to methods that enforce the expression of certain concepts or facts by activating the corresponding FFN neurons (see Section \ref{sec:related_work}), these methods could enforce the use of specific skills.

\section{Conclusion}
In this paper, we investigated the robustness of Prompt Tuning in relation to model activations, specifically focusing on the existence of skill neurons and their connection to adversarial robustness. Our findings revealed that Prompt Tuning yields prompts that are transferable between similar tasks but not robust to adversarial attacks. We identified skill neurons in both RoBERTa and T5, which were highly predictive and task-specific. Suppressing these skill neurons significantly impacted task performance, highlighting their importance. Interestingly, T5 demonstrated higher adversarial robustness than RoBERTa, and skill neurons in T5 exhibited stronger correlations between adversarial and non-adversarial data. This suggests a potential link between the activation of skill neurons on adversarial data and model robustness. Future research could explore methods to enhance model robustness by consistently activating relevant skill neurons. 

\section{Acknowledgements}
We are indebted to Yusheng Su for answering our questions about their prompt transferability experiments, and to Xiaozhi Wang for answering our questions about the skill neuron method. We would also like to thank Elia Bruni and Michael Rau for helpful discussions.

\nocite{*}
\section{Bibliographical References}\label{sec:reference}

\bibliographystyle{lrec-coling2024-natbib}
\bibliography{references.bib}


\appendix

\onecolumn

\section{Transferability}\label{app:transferability}

\begin{figure}[H]
    \centering
    \begin{subfigure}{0.49\textwidth}
        \includegraphics[width=\textwidth]{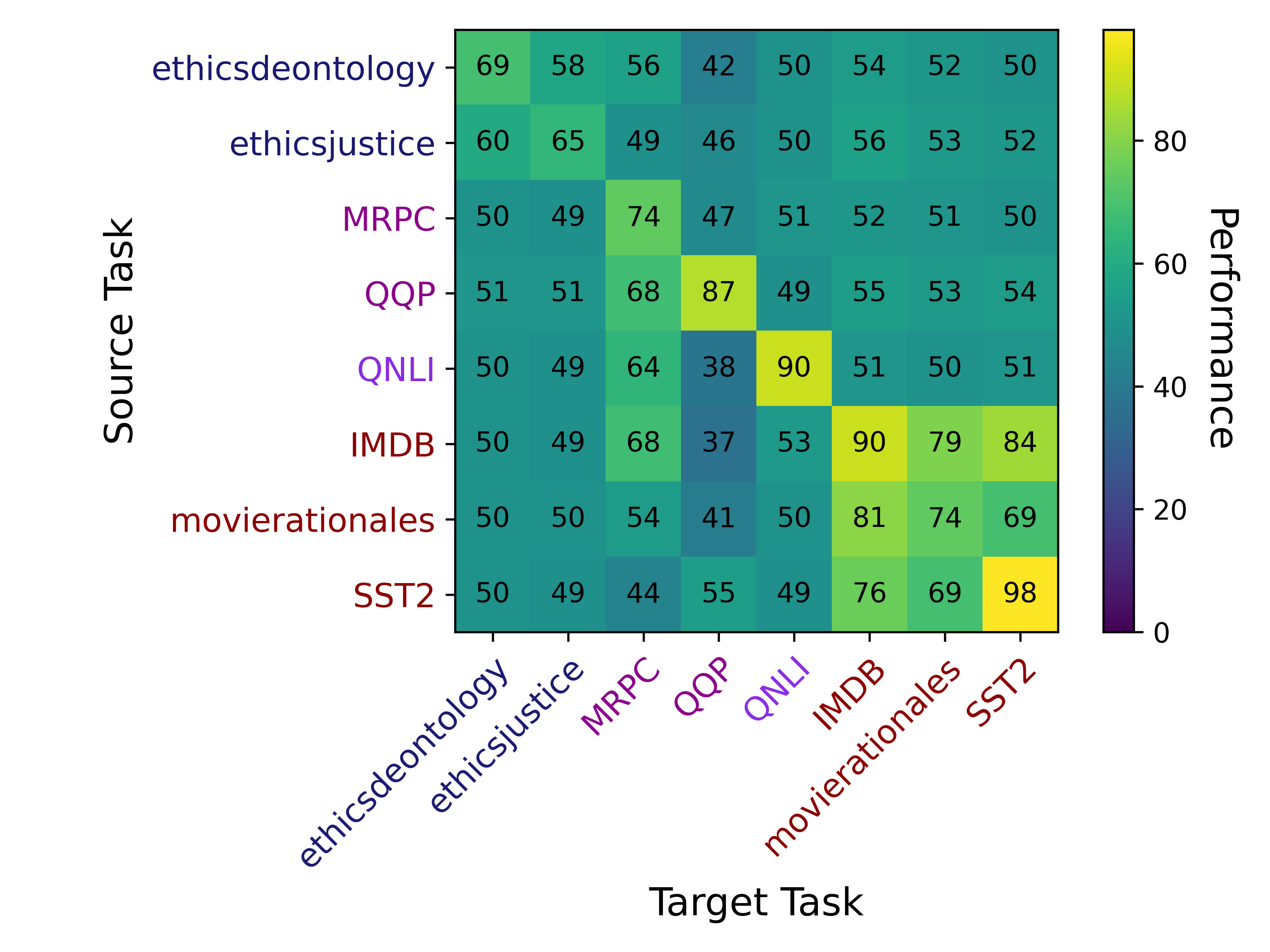} 
        \hspace{0.9cm}\caption{RoBERTa}
        \label{fig:subfig1_absolute_transfer}
    \end{subfigure}
    \begin{subfigure}{0.49\textwidth}
        \includegraphics[width=\textwidth]{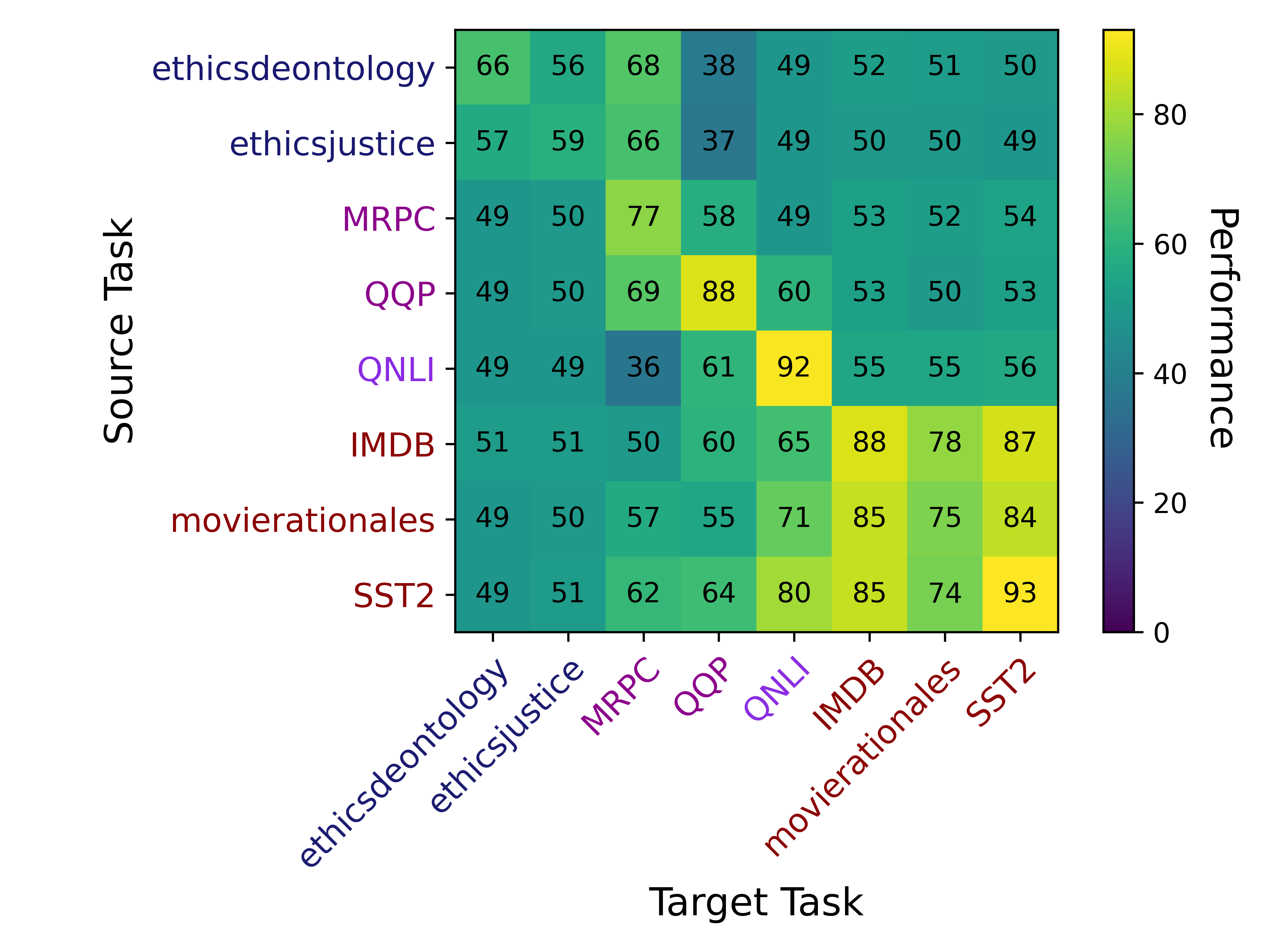} 
        \hspace{0.9cm}\caption{T5}
        \label{fig:subfig2_absolute_transfer}
    \end{subfigure}
    \caption{Prompt transferability. We calculate the accuracy when using the prompt for the source task on the target task.}
    \label{fig:absolute_transfer_performance}
\end{figure}

Figure \ref{fig:absolute_transfer_performance} shows the average absolute accuracy when evaluating the prompts tuned on a specific source dataset on a specific target dataset. Adversarial datasets are not included because we did not tune any prompts for these.  
Note that the matrices correspond to the matrices in Figure \ref{fig:transfer_performance} except that we report absolute instead of relative accuracies here. 

\section{Task-specificity (normalized)}\label{app:task_specificity}
\begin{figure}[H]
    \centering
    \begin{subfigure}[b]{0.49\textwidth}
        \includegraphics[width=\textwidth]{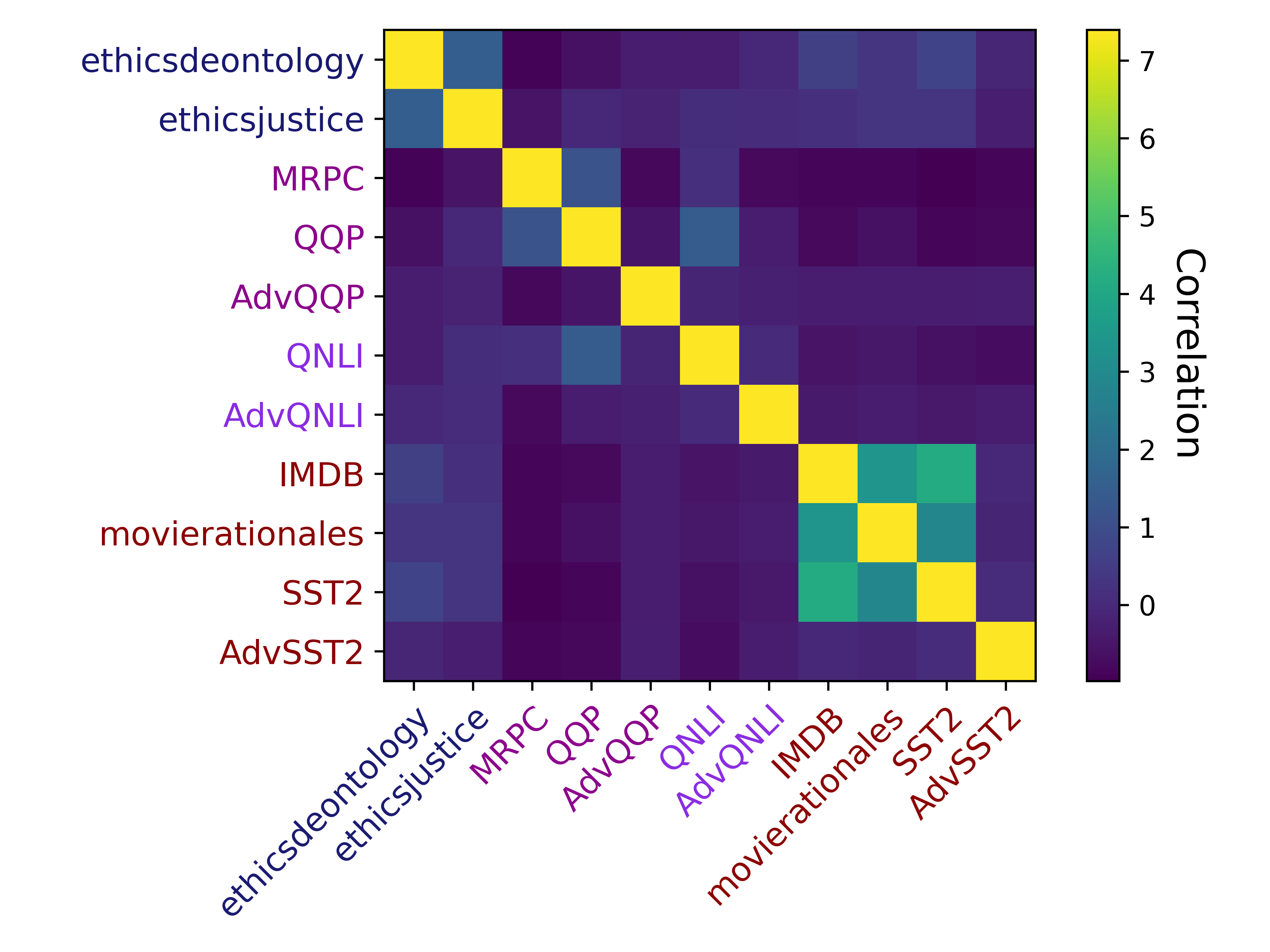} 
        \caption{RoBERTa}
        \label{fig:subfig1_specificity_normalized}
    \end{subfigure}
    \hfill
    \begin{subfigure}[b]{0.49\textwidth}
        \includegraphics[width=\textwidth]{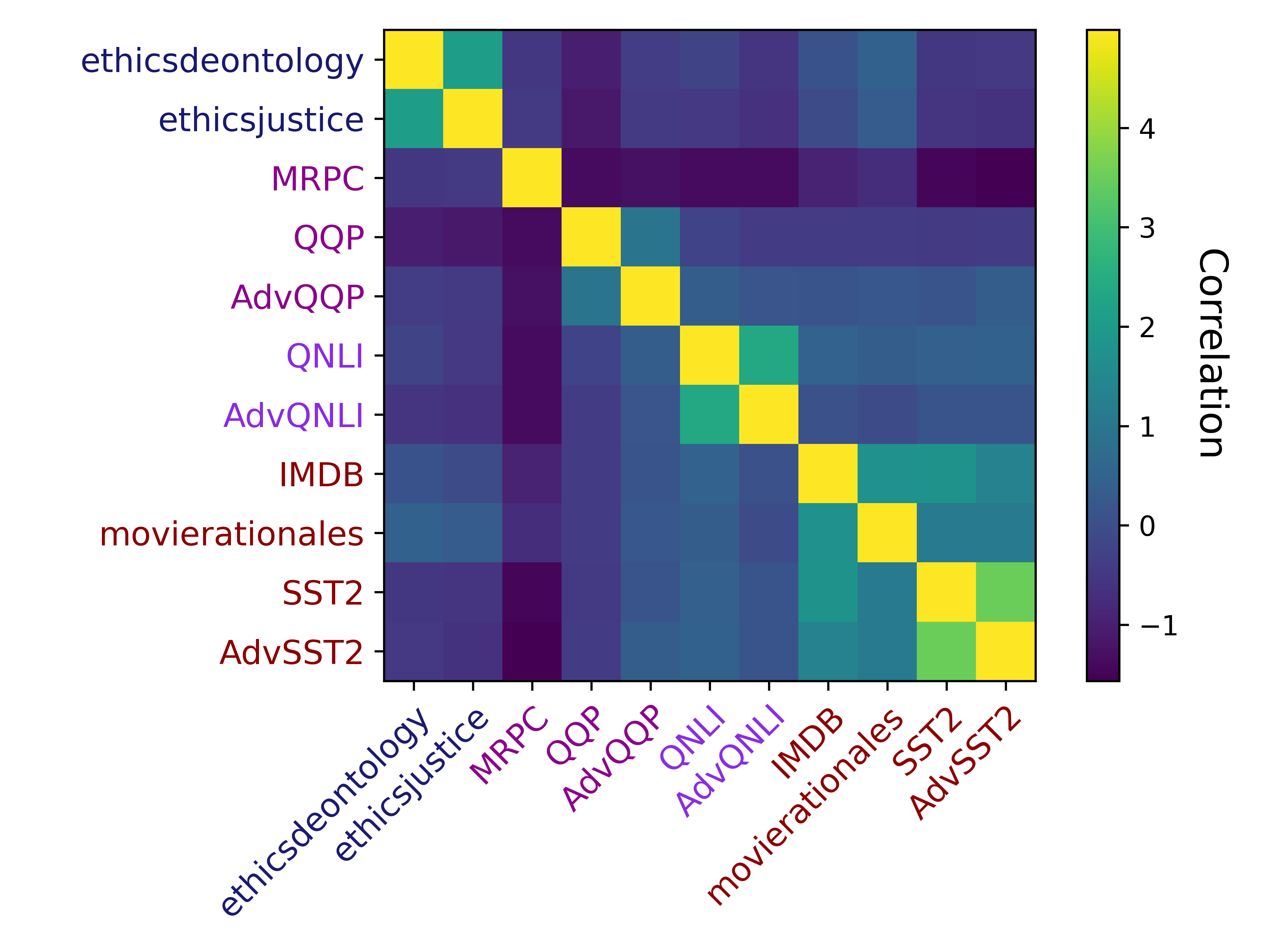} 
        \caption{T5}
        \label{fig:subfig2_specificity_normalized}
    \end{subfigure}
    \caption{Z-score normalized Spearman rank correlations of the neuron predictivities for different datasets.}
    \label{fig:task_specificity_normalized}
\end{figure}

The correlations between neuron predictivities for different datasets are generally higher for T5 than RoBERTa (see Figure \ref{fig:results_task_specificity}). 
We applied a Z-score normalization to the correlation values to account for this difference (see Figure \ref{fig:task_specificity_normalized}).
The normalization does not affect our conclusions:
Skill neurons in both T5 and RoBERTa are task-specific, and there is a strong correlation between neuron predictivities on adversarial and corresponding non-adversarial data for T5 but not RoBERTa.

\section{Neuron accuracies}\label{app:accuracies}

\begin{figure}[H]
    \centering
    \begin{subfigure}[b]{0.49\textwidth}
        \includegraphics[width=\textwidth]{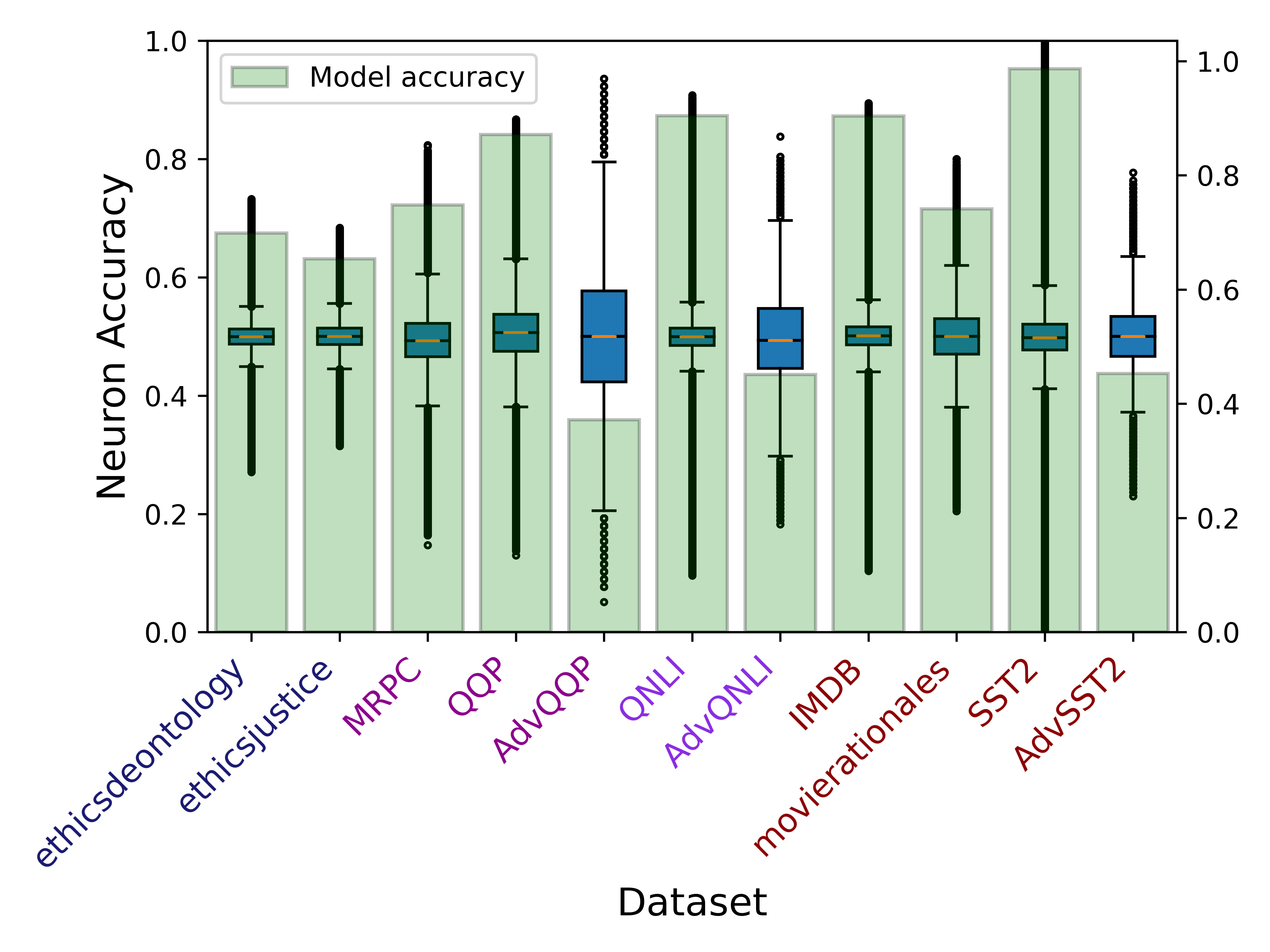}
        \caption{RoBERTa}
    \end{subfigure}
    \hfill 
    \begin{subfigure}[b]{0.49\textwidth}
        \includegraphics[width=\textwidth]{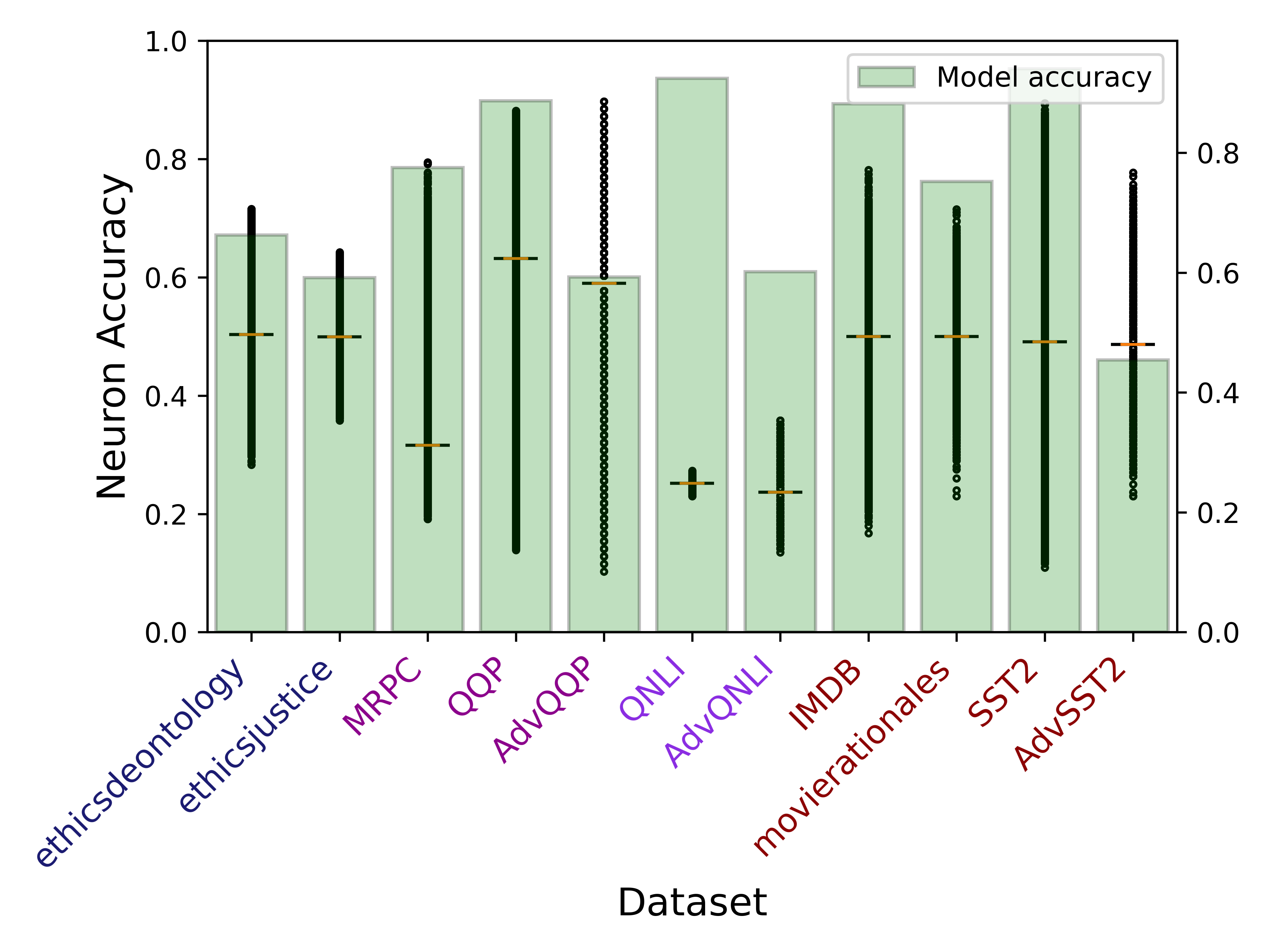}
        \caption{T5}
    \end{subfigure}
    \caption{Distribution of neuron accuracies (box plots) on top of model accuracy (bar plots).}
    \label{fig:results_boxplot_acc_acc_per_layer}
\end{figure}

Figure \ref{fig:results_boxplot_acc_acc_per_layer} shows the distributions of neuron accuracies (Equation \ref{eq:neuron_accuracies}) for both models and each dataset. For non-adversarial and adversarial datasets, neuron accuracies are excitatory and inhibitory. 
\texttt{(Adv)QNLI} poses an exception in that neuron activations are exclusively inhibitory.

\section{Suppression analysis}\label{app:suppression_analysis}

\begin{figure}[H]
    \centering
    \begin{subfigure}{0.32\textwidth}
        \centering
        \includegraphics[width=\textwidth]{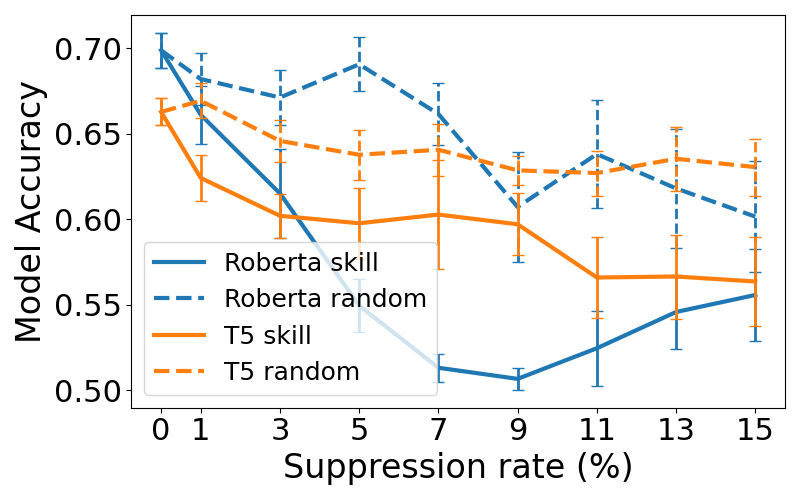}
        \caption{Ethics-Deontology}
    \end{subfigure}
    \begin{subfigure}{0.32\textwidth}
        \centering
        \includegraphics[width=\textwidth]{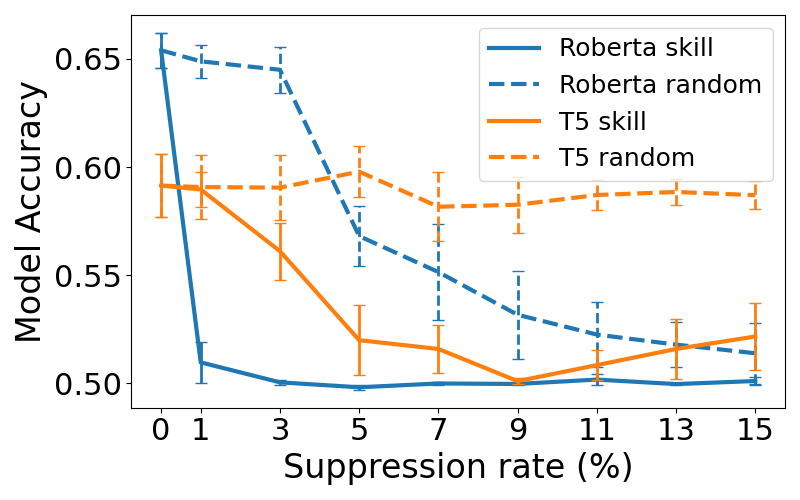}
        \caption{Ethics-Justice}
    \end{subfigure}
    \begin{subfigure}{0.32\textwidth}
        \centering
        \includegraphics[width=\textwidth]{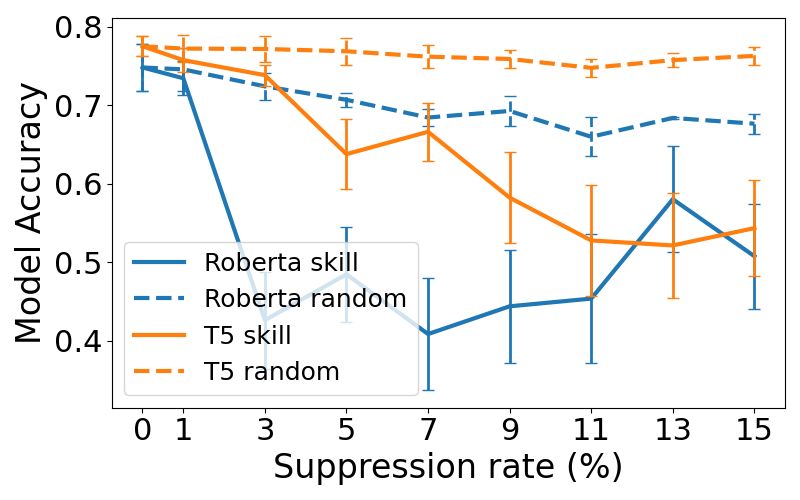}
        \caption{MRPC}
    \end{subfigure}
    \begin{subfigure}{0.32\textwidth}
        \centering
        \includegraphics[width=\textwidth]{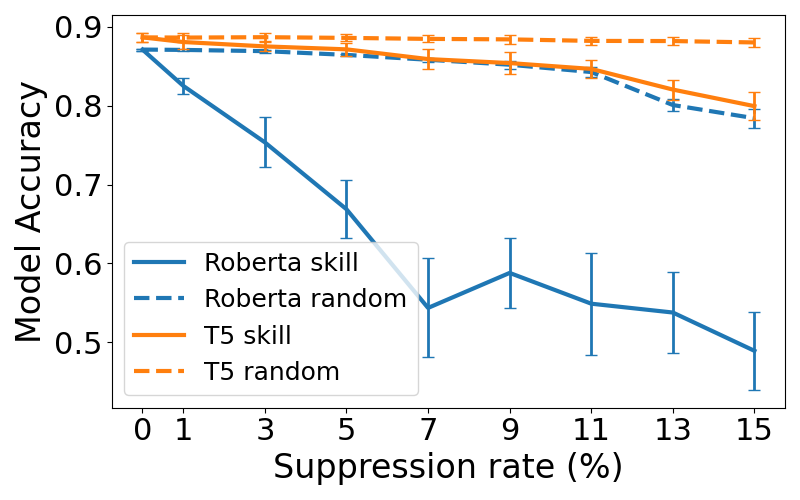}
        \caption{QQP}
    \end{subfigure}
    \begin{subfigure}{0.32\textwidth}
        \centering
        \includegraphics[width=\textwidth]{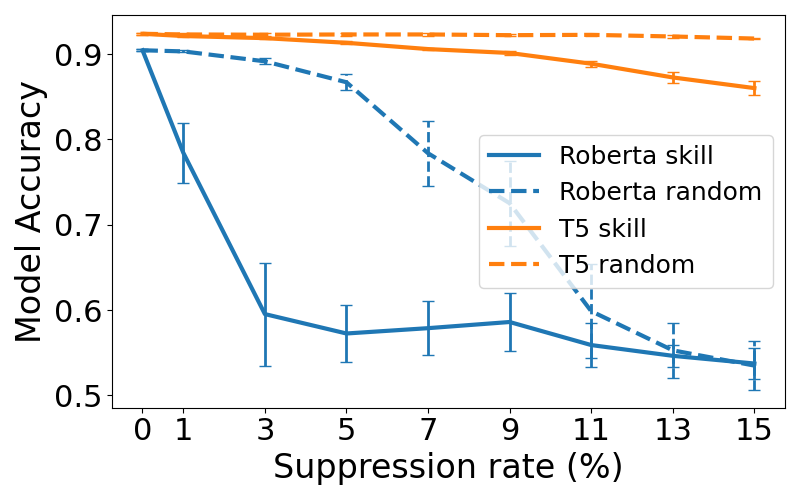}
        \caption{QNLI}
    \end{subfigure}
    \begin{subfigure}{0.32\textwidth}
        \centering
        \includegraphics[width=\textwidth]{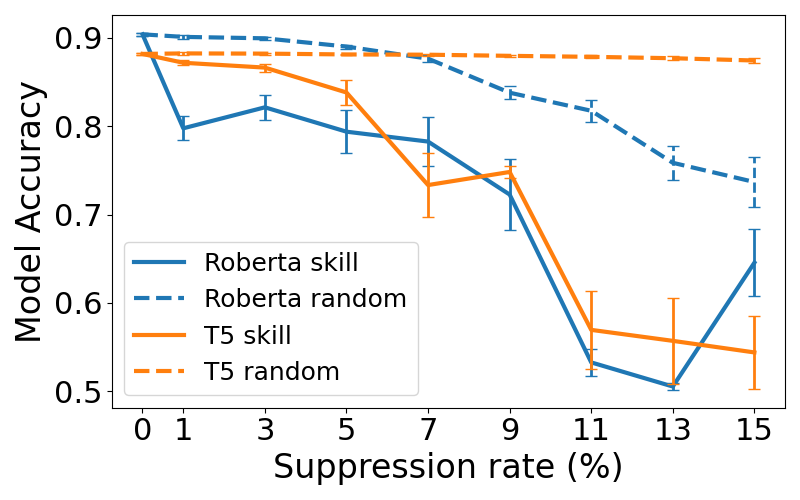}
        \caption{IMDB}
    \end{subfigure}
    \begin{subfigure}{0.32\textwidth}
        \centering
        \includegraphics[width=\textwidth]{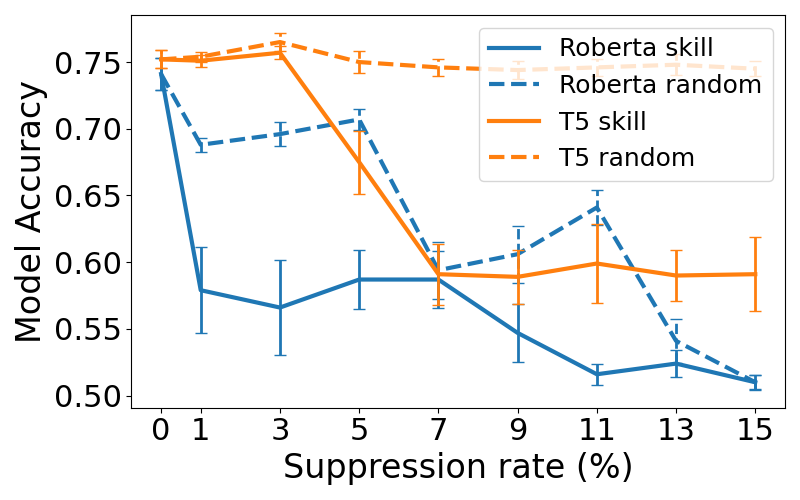}
        \caption{Movie Rationales}
    \end{subfigure}
    \begin{subfigure}{0.32\textwidth}
        \centering
        \includegraphics[width=\textwidth]{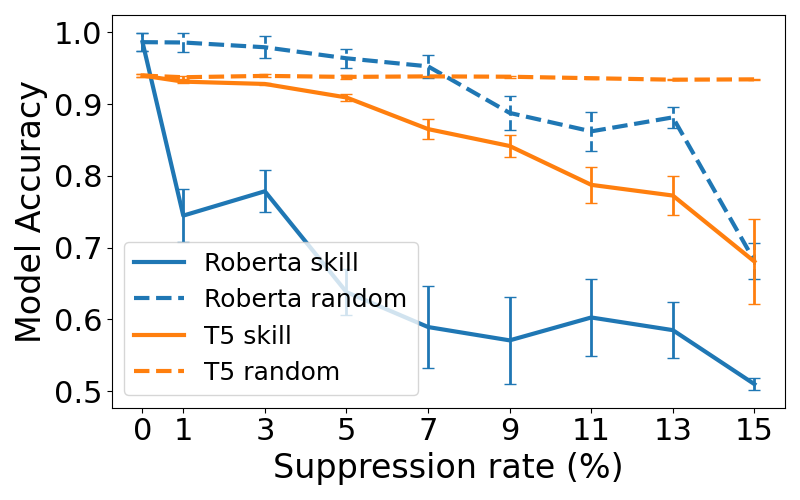}
        \caption{SST2}
    \end{subfigure}
    \caption{Model accuracies when neurons are suppressed. For each dataset, activations of the 0-15\% most predictive neurons (solid lines) or the same amount of randomly selected neurons (dashed lines) are set to zero.}
    \label{fig:suppression_analysis_non-adversarial}
\end{figure}

Figure \ref{fig:suppression_analysis_non-adversarial} shows the results of our suppression analysis for all non-adversarial datasets. 
Suppressing skill neurons consistently leads to a stronger decrease in accuracy than suppressing random neurons. In addition, accuracy tends to decrease with increasing suppression rates.
Suppressing random neurons hardly affects T5 but leads to a---sometimes strong---decrease in performance for RoBERTa. 

\end{document}